\documentclass[lettersize,journal]{IEEEtran}
\usepackage{amsmath,amsfonts}
\usepackage{algorithmic}
\usepackage{algorithm}
\usepackage{array}
\usepackage[caption=false,font=normalsize,labelfont=sf,textfont=sf]{subfig}
\usepackage{textcomp}
\usepackage{stfloats}
\usepackage{url}
\usepackage{verbatim}
\usepackage{graphicx}
\usepackage{cite}
\usepackage{adjustbox}
\usepackage{booktabs}
\usepackage{multirow}
\usepackage{colortbl}
\usepackage{tikz}
\usepackage{xcolor}
\usepackage{soul, color}

\definecolor{bestcolor}{rgb}{ .98,  .706,  .694}
\definecolor{secondcolor}{rgb}{1, .863, .769}
\definecolor{thirdcolor}{rgb}{ 1,  .976,  .89}

\begin{document}

\title{Remote Sensing Sparse-View 3D Gaussian Splatting \\ via Depth Image-Based Rendering }

\author{Jiaming Kang, Zhengxia Zou, and  Zhenwei Shi$^\star$ \\ Beihang University
% <-this % stops a space
}

\maketitle

\begin{abstract}
Remote sensing novel view synthesis under sparse observations remains challenging due to insufficient geometric constraints and limited cross-view supervision. Existing Neural Radiance Fields (NeRF) and 3D Gaussian Splatting (3DGS) methods are prone to overfitting and face challenges of depth ambiguities, missing cross-view information, and insufficient constraints in under-observed regions. To address these challenges, we propose DIBR-GS, a neural Gaussian Splatting framework that exploits Depth Image-Based Rendering (DIBR) to generate pseudo views for cross-view consistency supervision. Specifically, reliable geometric initialization is constructed by aligning monocular depth priors with sparse SfM reconstruction, and cross-view appearance priors are incorporated into neural Gaussian representations to enhance appearance modeling under sparse observations. Furthermore, we introduce a progressive DIBR-based pseudo-view supervision strategy to provide additional geometric and appearance constraints, enabling more complete reconstruction of weakly observed regions. In addition, a height-constrained anchor growth strategy is designed to suppress unreasonable Gaussian expansion. Experiments demonstrate that the proposed method achieves superior performance over existing approaches when training with only 3 input views. Compared with the previous best-performing method, it improves PSNR by 6.83 dB, with relative gains of 14\% in SSIM and 60\% in LPIPS, while maintaining competitive computational efficiency. Our code is available at \url{https://github.com/kanehub/DIBR-GS}
\end{abstract}


\section{Introduction}

Novel view synthesis aims to render images at arbitrary new viewpoints, given a set of images and their camera poses. In remote sensing, novel view synthesis provides a complementary way to enhance 3D scene understanding beyond conventional observations~\cite{ wu2022remote, 10529260, Mari_2022_CVPR, gao2025mp,wu2026progressive}. Consequently, it plays a crucial role in applications such as urban 3D reconstruction, disaster assessment, and environmental monitoring~\cite{10891840, 10654291,liu2025remoteVLM, 11505941}.

\begin{figure}[t]
    \centering
    \includegraphics[width=0.5\textwidth]{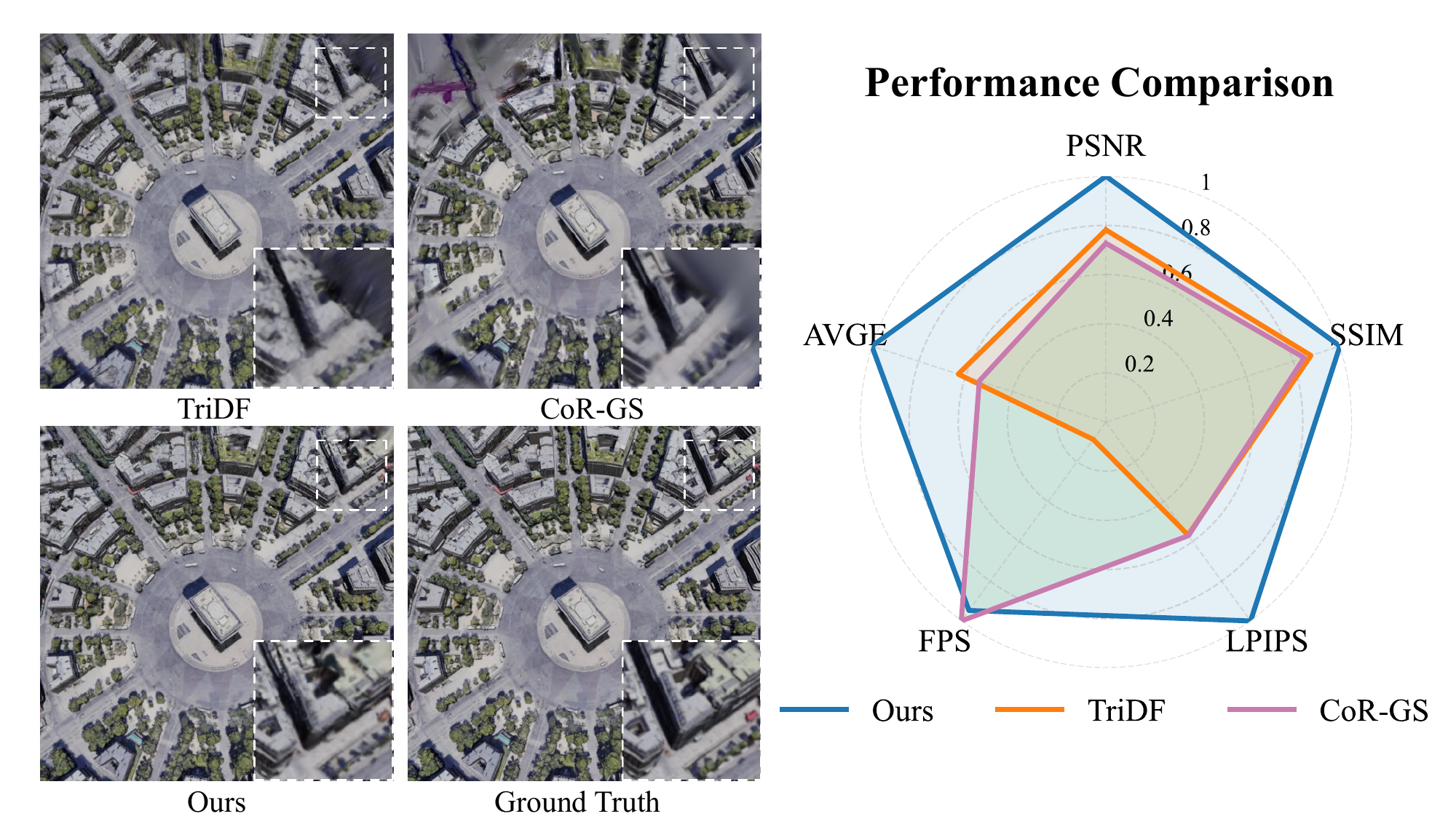}
    \caption{Visual and quantitative comparison on the LEVIR-NVS dataset with 3 input views. The proposed method produces more complete structures and finer details compared with existing approaches, especially in weakly observed regions. The radar chart summarizes the performance comparison in terms of reconstruction quality and rendering efficiency. For visualization, we report the negative LPIPS, AVGE, and $1+\log(\mathrm{FPS})$ values to enable unified comparison.}
    \label{fig:first}
\end{figure}

With the rapid development of deep learning, neural rendering has become the mainstream for novel view synthesis~\cite{yao2018mvsnet, tewari2020state}. Neural Radiance Fields (NeRF)~\cite{mildenhall2020nerf, xiangli2022bungeenerf, muller2022instant, 11265802} employs an implicit neural network to continuously model scene geometry and appearance, performing well in terms of view consistency and detail representation. However, its reliance on dense per-ray sampling incurs high computational costs. Conversely, 3D Gaussian Splatting (3DGS)~\cite{kerbl20233d,huang20242d, bao20253d, 11447307} has attracted widespread attention due to its real-time rendering capabilities. By optimizing explicit Gaussian primitives and leveraging tile-based rasterization, it significantly reduces both training and inference overhead while maintaining high-quality rendering results.

Despite these advances, existing neural rendering methods still heavily rely on dense multi-view observations. Under sparse-view conditions, the lack of sufficient multi-view constraints often leads to overfitting. This challenge becomes even more prominent in remote sensing scenes. Due to limitations such as satellite revisit cycles, drone flight paths, and occlusions, it is difficult to capture dense multi-view images. In extreme cases, only 3 to 5 viewpoints are available, which is far fewer than the hundreds of densely sampled multi-view inputs typically used. This observation scarcity leaves large regions of the viewpoint space weakly constrained, resulting in ambiguous geometry, incomplete structures, and inconsistent appearance during novel-view synthesis.

For sparse-view inputs, NeRF-based methods typically rely on regularization~\cite{niemeyer2022regnerf, yang2023freenerf, somraj2023vip}, semantic priors~\cite{jain2021putting, guo2025taco}, or depth supervision~\cite{deng2022depth,roessle2022dense}. However, these strategies are mainly developed for object-centric scenes with relatively diverse viewpoint coverage, and their effectiveness remains limited for remote sensing scenes, which contain complex land-cover structures observed from constrained overhead viewpoints. Meanwhile, 3DGS-based approaches attempt to enhance sparse-view performance by incorporating depth priors~\cite{zhu2024fsgs,Li_2024_CVPR,chung2024depth}, co-regularization~\cite{zhang2024cor}, or structural pruning~\cite{Park_2025_CVPR}. Nevertheless, depth ambiguities in remote sensing images can compromise reconstruction accuracy, and heuristic optimization strategies struggle to capture complex scene geometry and maintain cross-view consistency.

To this end, we identify the fundamental challenge of sparse-view novel view synthesis in remote sensing as how to effectively exploit various prior information, including geometry, appearance, and structural relationships, from limited observations. Existing methods are largely constrained by the unreliability and insufficient utilization of such priors, which can be manifested in three aspects. First, geometric priors derived from monocular depth estimation suffer from scale ambiguity and structural distortions in remote sensing images, making them unreliable for accurate supervision or initialization. Second, existing approaches lack effective mechanisms to jointly exploit appearance and geometric cues across multiple views, leading to inconsistent reconstruction of scene structures and textures. Third, the lack of effective supervision for unobserved regions, where observations are limited due to occlusions or viewpoint constraints, often leads to incomplete geometry recovery and degraded reconstruction quality.

To address the above challenges, we propose DIBR-GS, a geometry-guided neural Gaussian splatting framework. It enhances sparse-view novel view synthesis in remote sensing scenes by unifying initialization, feature representation, and unobserved-view constraints within a depth image-based rendering (DIBR)~\cite{sun2010overview} framework, which effectively exploits cross-view geometric and appearance priors. Specifically, we first design a point cloud initialization strategy tailored for monocular depth estimation. An affine transformation is introduced to alleviate scale ambiguity of monocular depth, while multi-view fusion is employed to reduce geometric misalignment and obtain reliable initial anchors. Then, following the image-based rendering (IBR)~\cite{9578424} framework, cross-view geometric and appearance features are incorporated into the feature representation of neural Gaussians, improving the utilization of limited prior information. More importantly, we introduce a progressive pseudo-view synthesis strategy via DIBR to propagate depth priors from input images to unobserved viewpoints, which effectively promotes multi-view consistency during Gaussian optimization. To further improve robustness for near-nadir remote sensing scenes, we incorporate a height-constrained anchor growth strategy based on aerial scene priors to suppress floating artifacts.

Ultimately, our proposed method achieves high-fidelity novel view synthesis from sparse-view inputs in remote sensing scenes while preserving real-time rendering capability. Fig.~\ref{fig:first} presents the visual and quantitative comparison results.

In summary, our main contributions are:

\begin{itemize}
  \item [1)] 
  We propose DIBR-GS, which integrates depth-based image rendering with Gaussian Splatting to effectively exploit geometric and appearance priors for remote sensing sparse view synthesis.
  \item [2)]
  We introduce a progressive pseudo view synthesis strategy that propagates depth priors from observed views to unobserved viewpoints, providing additional multi-view consistency constraints during Gaussian optimization. This strategy effectively alleviates incomplete geometry and structural degradation in under-observed regions.
  \item [3)]
  Experiments on the LEVIR-NVS~\cite{wu2022remote} dataset demonstrate that the proposed method consistently outperforms state-of-the-art approaches for remote sensing sparse view synthesis, achieving a favorable balance between novel-view rendering quality and computational efficiency.
\end{itemize}

\section{Related Work}

\subsection{Neural Scene Representations}
Neural scene representation refers to using neural networks to approximate the surface or volumetric representation function, and integrating classical rendering principles to achieve a continuous and compact 3D representation. Based on the underlying structural form, these methods can be categorized into implicit, explicit, and hybrid representations~\cite{tewari2020state}. Neural Radiance Fields~\cite{mildenhall2020nerf,barron2021mip} serves as a classical implicit approach, which maps spatial coordinates to color and volume density via a multilayer perceptron, enabling high-fidelity novel view synthesis. However, the training efficiency is limited due to the dense sampling and per-ray inference. Hybrid representations~\cite{liu2020neural, Fridovich-Keil_2022_CVPR, chen2022tensorf} incorporate explicit geometry to reduce computational costs, while maintaining modeling capabilities through lightweight decoding networks. Plenoxels~\cite{Fridovich-Keil_2022_CVPR} store spherical harmonic coefficients and volume density within a sparse voxel grid to enable efficient queries. Point-NeRF~\cite{xu2022point} embeds neural features within point clouds, which accelerates convergence and provides explicit editing capabilities. Alternative methods encode scene information using multi-plane images~\cite{richard2020single,li2021mine} or triplane representations~\cite{hu2023tri} to avoid the redundancy associated with volumetric approaches. 

Recently, explicit representations such as 3D Gaussian Splatting~\cite{kerbl20233d} represent scenes with anisotropic Gaussian primitives and rasterization, achieving high-fidelity reconstruction and real-time rendering. 2DGS~\cite{huang20242d} introduces surface Gaussian primitives and constrains them to align closely with object surfaces to improve geometric accuracy. Scaffold-GS~\cite{lu2024scaffold} introduces an anchor-based hierarchical structure that dynamically generates and organizes neural Gaussians utilizing a sparse voxel grid, addressing issues of structural redundancy and loose organization. Although these approaches resolve the trade-offs among rendering speed, geometric quality, and storage efficiency, the training process still requires massive amounts of multi-view images to ensure convergence.

\subsection{Novel View Synthesis with Sparse Views}

Novel view synthesis aims to render images from unobserved viewpoints based on input images and their corresponding poses. Traditional methods require hundreds of views to provide dense supervision, which is impractical for remote sensing scenarios because data acquisition is often constrained by flight path or fuel and collected images are typically sparse. Current research primarily improves sparse-view performance on object-level and synthetic scenes. Early methods introduce regularization constraints to mitigate overfitting in neural radiance fields under sparse views~\cite{yang2023freenerf,jain2021putting,niemeyer2022regnerf,somraj2023vip}. For instance, RegNeRF~\cite{niemeyer2022regnerf} employs local depth smoothness to eliminate irregular artifacts. DietNeRF~\cite{jain2021putting} enforces cross-view semantic consistency. FreeNeRF~\cite{yang2023freenerf} designs a frequency regularization strategy to achieve progressive learning from low to high frequencies. DS-NeRF~\cite{deng2022depth} utilizes sparse point clouds as depth supervision, providing effective explicit constraints for scene geometry. Other approaches~\cite{yu2021pixelnerf, chen2021mvsnerf,9578424} explore cross-scene pretraining strategies to learn general scene representations and view synthesis priors from large-scale data, enhancing few-shot performance and generalization.

3DGS also suffers performance degradation under sparse viewpoints. To address this, some methods incorporate depth priors as auxiliary supervision~\cite{zhu2024fsgs, Li_2024_CVPR, chung2024depth}. FSGS~\cite{zhu2024fsgs} guides the initialization and densification of Gaussian primitives using monocular depth. By distributing Gaussian points more appropriately in space, FSGS compensates for holes in sparse initial SfM point clouds, enabling coherent scene reconstruction even with very few samples. Similarly, DNGaussian~\cite{Li_2024_CVPR} employs hard and soft depth regularization losses to enforce consistency between predicted depth and monocular depth, resolving common depth ambiguities under sparse views. Other methods focus on regularization strategies to suppress artifacts and optimize Gaussian distributions~\cite{Park_2025_CVPR,zhang2024cor,avinash2024coherent,zheng2025nexusgs}. DropGaussian~\cite{Park_2025_CVPR} identifies and removes redundant primitives that cause visual artifacts by analyzing cumulative opacity and spatial distribution, resulting in a more compact scene representation. CoR-GS~\cite{zhang2024cor} jointly trains two Gaussian fields and utilizes their inconsistencies to suppress inaccurate reconstructions, improving view synthesis quality. Nevertheless, these methods still face challenges in remote sensing scenarios. Monocular depth exhibits inherent scale ambiguity, making depth constraints unreliable. Moreover, manually designed heuristic constraints struggle to handle the complex structures present in remote sensing scenes.

\begin{figure*}
    \centering
    \includegraphics[width=1.0\textwidth]{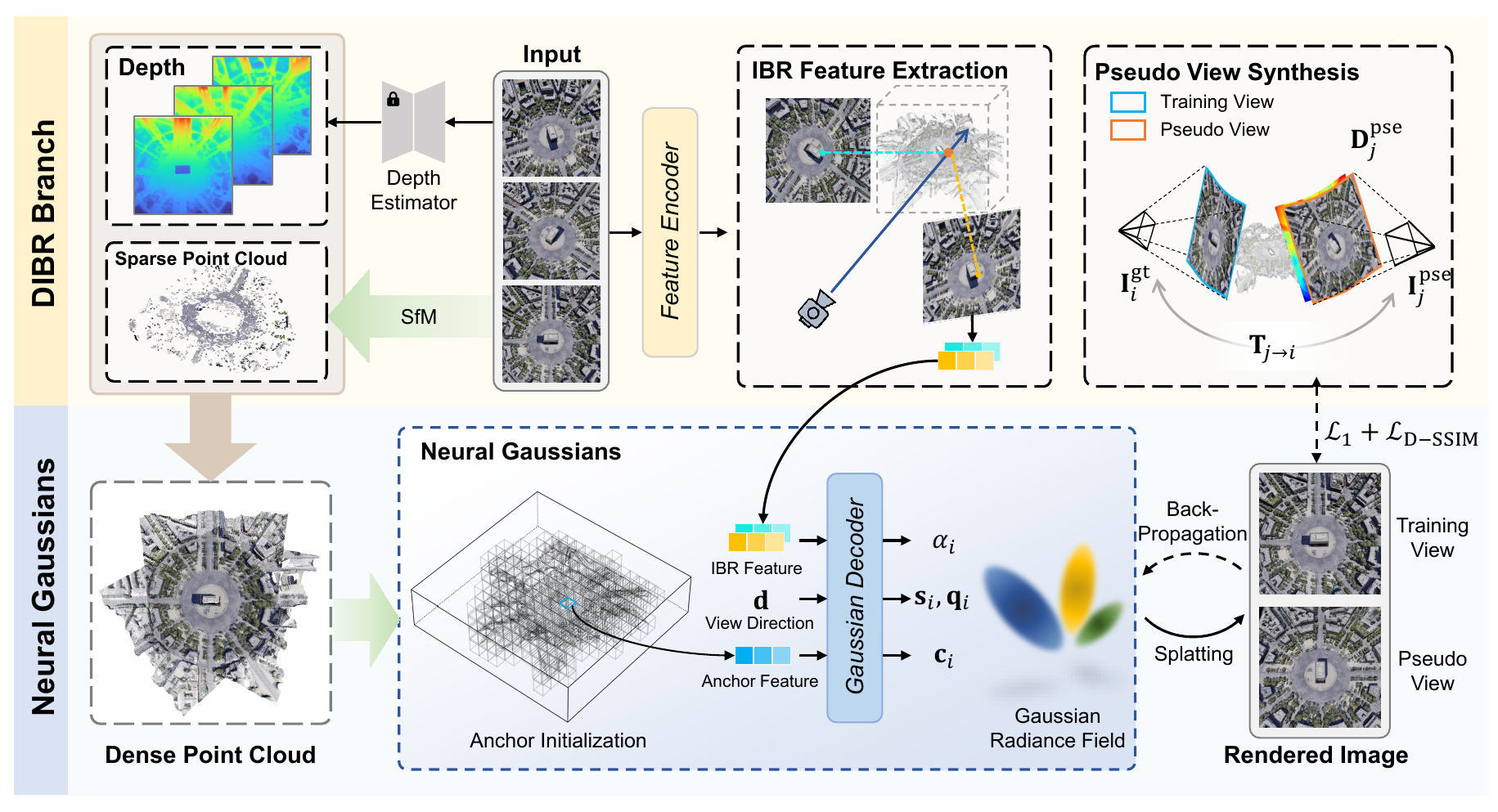}
    \caption{Overview of the proposed DIBR-GS framework. The framework consists of a DIBR branch and a neural Gaussian branch. In the DIBR branch, monocular depth priors are aligned with sparse SfM reconstruction through affine transformation and multi-view fusion to construct a reliable geometric initialization. Cross-view appearance features are extracted following the image-based rendering paradigm, and progressive DIBR is further employed to generate pseudo views that provide additional supervision. In the neural Gaussian branch, Gaussian anchors are initialized from the dense point cloud, while the anchor features and cross-view appearance features are adaptively integrated to predict Gaussian attributes. The rendered training views and pseudo views are jointly optimized, enabling more complete and consistent reconstruction under sparse-view remote sensing scenes.}
    \label{fig:overview}
\end{figure*}

\section{Method}

We propose DIBR-GS, a geometry-guided neural Gaussian framework. Inspired by the DIBR, we explore how to effectively leverage reliable geometric and appearance priors through three key stages: geometry initialization, feature representation, and optimization constraints. The overall framework is illustrated in Fig.~\ref{fig:overview}. First, monocular depth priors are aligned with SfM points and fused across multiple views to construct reliable initial Gaussian anchors, reducing geometric ambiguity under sparse inputs. Then, cross-view reference features are incorporated into neural Gaussian representations to provide complementary appearance cues for sparse-view optimization. Finally, unlike conventional DIBR that aims at producing visually complete images, we utilize progressive DIBR to establish additional supervision in unobserved regions and viewpoints. The generated pseudo views implicitly incorporate geometric information derived from depth priors and enhance multi-view consistency during optimization. Besides, considering the inherent height distribution prior of remote sensing scenes, we further introduce a height-constrained anchor growth strategy to constrain unreasonable Gaussian expansion.

\subsection{Preliminary}

\subsubsection{3DGS}

3DGS~\cite{kerbl20233d} represents a scene using a set of anisotropic Gaussian primitives. Given the mean $\boldsymbol{\mu} \in \mathbb{R}^3$ and covariance matrix $\mathbf{\Sigma} \in \mathbb{R}^{3\times3}$, each Gaussian is defined as
\begin{equation}
    G(\mathbf{x})=
    e^{-\frac{1}{2}(\mathbf{x}-\boldsymbol{\mu})^T
    \mathbf{\Sigma}^{-1}
    (\mathbf{x}-\boldsymbol{\mu})}.
\end{equation}

To ensure positive semi-definiteness, the covariance matrix is parameterized as
$\mathbf{\Sigma}=\mathbf{R}\mathbf{S}\mathbf{S}^T\mathbf{R}^T$, where $\mathbf{R}$ and $\mathbf{S}$ are represented by a quaternion $\mathbf{q}\in\mathbb{R}^4$ and a scaling vector $\mathbf{s}\in\mathbb{R}^3$, respectively. Each Gaussian is further associated with a color $\mathbf{c}$ and opacity $\alpha$.

During rendering, 3D Gaussians are projected onto the image plane and blended in depth order through differentiable rasterization. The pixel color is computed by
\begin{equation}
    C=\sum_{i=1}^{N}\mathbf{c}_i\sigma_i
    \prod_{j=1}^{i-1}(1-\sigma_j),
    \quad
    \sigma_i=\alpha_iG_i^{\prime}(\mathbf{x}),
\end{equation}

\subsubsection{Neural Gaussian}

Scaffold-GS~\cite{lu2024scaffold} introduces anchor-based neural Gaussians for compact scene representation. Each anchor controls $k$ Gaussians, whose centers are generated by
\begin{equation}
    \{\boldsymbol{\mu}_1,\ldots,\boldsymbol{\mu}_{k}\}
    =
    \mathbf{x}_v+
    \{\mathcal{O}_1,\ldots,\mathcal{O}_{k}\}\cdot l_v,
\end{equation}
where $\mathbf{x}_v$ is the anchor position, $l_v$ is a scaling factor, and $\mathcal{O}_k\in\mathbb{R}^3$ denotes the learned offset.

The Gaussian attributes $\{\alpha_i,\mathbf{q}_i,\mathbf{s}_i,\mathbf{c}_i\}$ are decoded from the anchor feature $\mathbf{f}_s\in\mathbb{R}^{32}$, the viewing direction $\mathbf{d}_{vc}$, and the camera-relative distance $\delta_{vc}$ using lightweight MLPs. For example, opacity is predicted as
\begin{equation}
    \{\alpha_1,\ldots,\alpha_k\}
    =
    F_{\alpha}(\mathbf{f}_s,\delta_{vc},\mathbf{d}_{vc}),
\end{equation}
while quaternion, scale, and color are generated by $F_q$, $F_s$, and $F_c$, respectively.

During training, anchors are adaptively grown according to Gaussian gradients and pruned when their opacity remains low, enabling efficient coverage of scene structures.

\subsection{Initialization with Dense Depth Priors}

Point cloud initialization plays an important role in sparse-view 3D Gaussian optimization, as inaccurate geometric initialization may lead to incomplete structures and unstable optimization. However, SfM reconstruction from extremely sparse views usually produces insufficient points, limiting the ability to capture scene structures. To alleviate this issue, we exploit monocular depth estimation as an auxiliary geometric prior and align it with SfM geometry to construct a denser and more reliable initialization. Given an input image $\mathbf{I}$, the original monocular depth map can be formulated as $\mathbf{D} = F_\theta(\mathbf{I}).$ 
Since monocular depth estimation only provides relative depth, we align it with sparse SfM points through an affine transformation:

\begin{equation}
    \hat{\mathbf{D}}  = a\cdot \mathbf{D} + b .
\end{equation}

The scaling and translation coefficients a, b are solved using the least squares method:

\begin{equation}
    \hat{a},\hat{b}=\arg\min_{a,b}\sum_{p\in \Omega_{\mathrm{sfm}}}\left(\hat{\mathbf{D}}(p;a,b)-\mathbf{D}_{\mathrm{sfm}}(p)\right)^2,
\end{equation}

\noindent where $p = (u,v)$ represents the projected coordinates of the point clouds on the image, and $\mathbf{D}_{\mathrm{sfm}}$ is the depth of the sparse SfM point clouds. Subsequently, we use the depth map to back-project image pixels into the 3D space of the world coordinate system:

\begin{equation}
    \mathbf{P}=\mathbf{R}_i\left(\hat{\mathbf{D}}_{i}(u,v)\mathbf{K}_i^{-1}\mathbf{p}\right)+\mathbf{t}_i,
\end{equation}

\noindent where $\mathbf{P} = (x,y,z,1)^\top$ represents the homogeneous coordinates in the world coordinate system, $\mathbf{K}_i$ denotes the intrinsics of view $i$, $\mathbf{R}_i$ and $\mathbf{t}_i$ represent the rotation and translation matrix in the extrinsics, and $\mathbf{p} = (u,v,1)^\top$ represents the homogeneous pixel coordinates.

After back-projecting depth maps from multiple views, we fuse the generated points to obtain an initial dense representation. However, due to inherent errors in monocular depth estimation, the fused point cloud often contains numerous mismatches. In remote sensing scenarios, such errors typically manifest as layered stacking of point clouds at different heights, which clearly contradicts reality. For near-nadir remote sensing scenes, dominant structures can often be approximated by a single surface distribution along the vertical direction. We therefore exploit this domain prior to identify obvious depth outliers while preserving the major scene structure.

Leveraging this prior knowledge, we can effectively identify and remove inaccurate point clouds.

We first partition the point clouds into grids on the $XY$ plane at a fixed resolution, with each grid denoted as $G_k$. The height set of point clouds within each grid is denoted as:

\begin{equation}
    Z_k = \{z_i|(x_i,y_i)\in G_k\}
\end{equation}

We perform a dip test~\cite{hartigan1985dip} on each grid. If the height distribution significantly deviates from the hypothesis of unimodality, we consider that the region may contain multiple vertically overlapping surfaces and unreliable points. To effectively evaluate the overlapping relationships, we construct a KD-tree~\cite{bentley1975multidimensional} within the grid to identify neighboring point pairs. Two points $P_i$ and $P_j$ are considered overlapping if they satisfy the following criterion:

\begin{equation}
    \mathbb{I}_{\mathrm{ovlp}}(P_i,P_j)=
\begin{cases}
1, & {\text{if}}\  \|(x_i,y_i)-(x_j,y_j)\|_2<\varepsilon_{xy} \\ & \wedge  |z_i-z_j|>\varepsilon_z \\
0, & \mathrm{otherwise,} 
\end{cases}
\end{equation}

\noindent where $\varepsilon_{xy}$ and $\varepsilon_z$ represent the thresholds for horizontal and vertical distances, respectively.

Previous studies~\cite{song2026dgs} have revealed that sparse-view Gaussian optimization often suffers from a typical failure mode where Gaussians tend to accumulate toward the cameras, resulting in overfitting to input views. Therefore, when multiple redundant surfaces are generated during depth fusion, we retain the points with lower height values as initialization anchors. For near-nadir remote sensing scenes, lower-height points generally correspond to surfaces farther from the camera, which provides a more stable initialization and mitigates the camera-oriented drift of optimized Gaussians.

\subsection{IBR-Based Neural Gaussian Representation}

To enhance modeling under sparse observations, we incorporate cross-view reference features into neural Gaussian representations. The proposed representation exploits complementary appearance cues from multiple observations following an image-based rendering paradigm. However, due to occlusions and inaccurate matches in sparse views, reference features may contain unreliable information. Therefore, we further introduce an adaptive modulation mechanism to control the contribution of cross-view features during Gaussian optimization.

\subsubsection{IBR Feature Extraction}

Given an input image $\mathbf{I}$, 
we adopt a pretrained DINO~\cite{caron2021emerging} encoder to extract the local features pyramid $\{\mathbf{F}_r^\ell\}_{\ell=1}^L$ and the ViT tokens $\mathbf{F}_t$ that contain global semantics and long-range dependencies. We subsequently introduce a learnable projection module $\mathcal{P}$ to map both into the same dimension and fuse them into a single feature map, thereby modeling the complementary relationship between local texture and global semantics within a unified feature space:

\begin{align}
\tilde{\mathbf{F}}_r^\ell&=\operatorname{Up}(\mathcal{P}_\ell(\mathbf{F}_r^\ell)),\\
\tilde{\mathbf{F}}_t&=\operatorname{Broadcast}(\mathcal{P}_t(\mathbf{F}_t)),\\
\mathbf{F}_v &= \tilde{\mathbf{F}}_t + \sum_{\ell=1}^L \tilde{\mathbf{F}}_r^\ell ,
\end{align}

\noindent where $\operatorname{Up}(\cdot)$ represents upsampling and $\operatorname{Broadcast}(\cdot)$ expands patch-level features to pixel space by repeating.

For anchor point with center $\boldsymbol{\mu}_w=(x_w,y_w,z_w)^\top$, its projection coordinate in the $i$-th view is given as:

\begin{equation}
\begin{aligned}
    \boldsymbol{\mu}_c &= \mathbf{E}_i [\boldsymbol{\mu}_w ;1], \\
    \mathbf{p}_i &= \frac{1}{z_c}\mathbf{K}_i\boldsymbol{\mu}_c, 
\end{aligned}
\end{equation}

\noindent where $\boldsymbol{\mu}_c = (x_c,y_c,z_c)^\top$ and $\mathbf{E}_i$ represents the extrinsic matrix from world to camera. Then we extract the feature vector from the feature map according to the projection coordinate:

\begin{equation}
    \mathbf{f}_{v,i}=\Psi\left(\mathbf{F}_v,\mathbf{p}_i\right),
\end{equation}

\noindent where  $\Psi(\cdot)$  denotes bilinear sampling. By aggregating the sampled vectors across all views, we derive the final IBR feature for the anchor:

\begin{equation}
    \mathbf{f}_v=\frac{1}{|\mathcal{V}|}\sum_{i\in\mathcal{V}}\mathbf{f}_{v,i}.
\end{equation}

\subsubsection{Adaptive Feature Fusion}

Given that the IBR features under sparse views are susceptible to occlusion or matching errors, the direct fusion of reference features is prone to introducing noise. To address this, we design a modulation network $\mathcal{E}_\mathrm{mod}$ that 
adaptively predicts channel-wise modulation weights to regulate the contribution of cross-view features:

\begin{equation}
    \mathbf{g} = \mathcal{E}_\mathrm{mod}(\mathbf{f}_v).
\end{equation}

The fused feature of the anchor point is expressed as:

\begin{equation}
    \hat{\mathbf{f}}_s = \mathbf{f}_s + \mathbf{g} \odot \mathcal{E}_\mathrm{ref}(\mathbf{f}_v),
\end{equation}

\noindent where $\mathbf{f}_s$ denotes the original neural Gaussian feature, $\mathcal{E}_\mathrm{ref}$ represents the feature projection network, and $\odot$ indicates channel-wise multiplication.
Through the adaptive feature fusion strategy, information derived from IBR is fused into the neural Gaussian representation in a controlled and stable manner, reducing the influence of unreliable reference features while preserving useful prior information for sparse-view reconstruction.

\subsection{Height-constrained Anchor Growth}

A typical failure mode in sparse-view novel view synthesis is the emergence of floating artifacts around cameras, which mainly originate from regions with insufficient observation constraints~\cite{yang2023freenerf}. Due to the limited multi-view overlap, Gaussian optimization may explain uncertain regions by increasing primitive density near the observed viewpoints, resulting in unstable geometry and noticeable artifacts. Although depth-based regularization strategies~\cite{zhu2024fsgs, Li_2024_CVPR} have been explored to alleviate this issue, they usually require reliable depth supervision. In this work, we exploit the inherent characteristics of remote sensing scenes and introduce a lightweight geometry-aware regularization strategy to constrain unreasonable anchor growth.

\begin{figure}[t]
    \centering
    \includegraphics[width=0.5\textwidth]{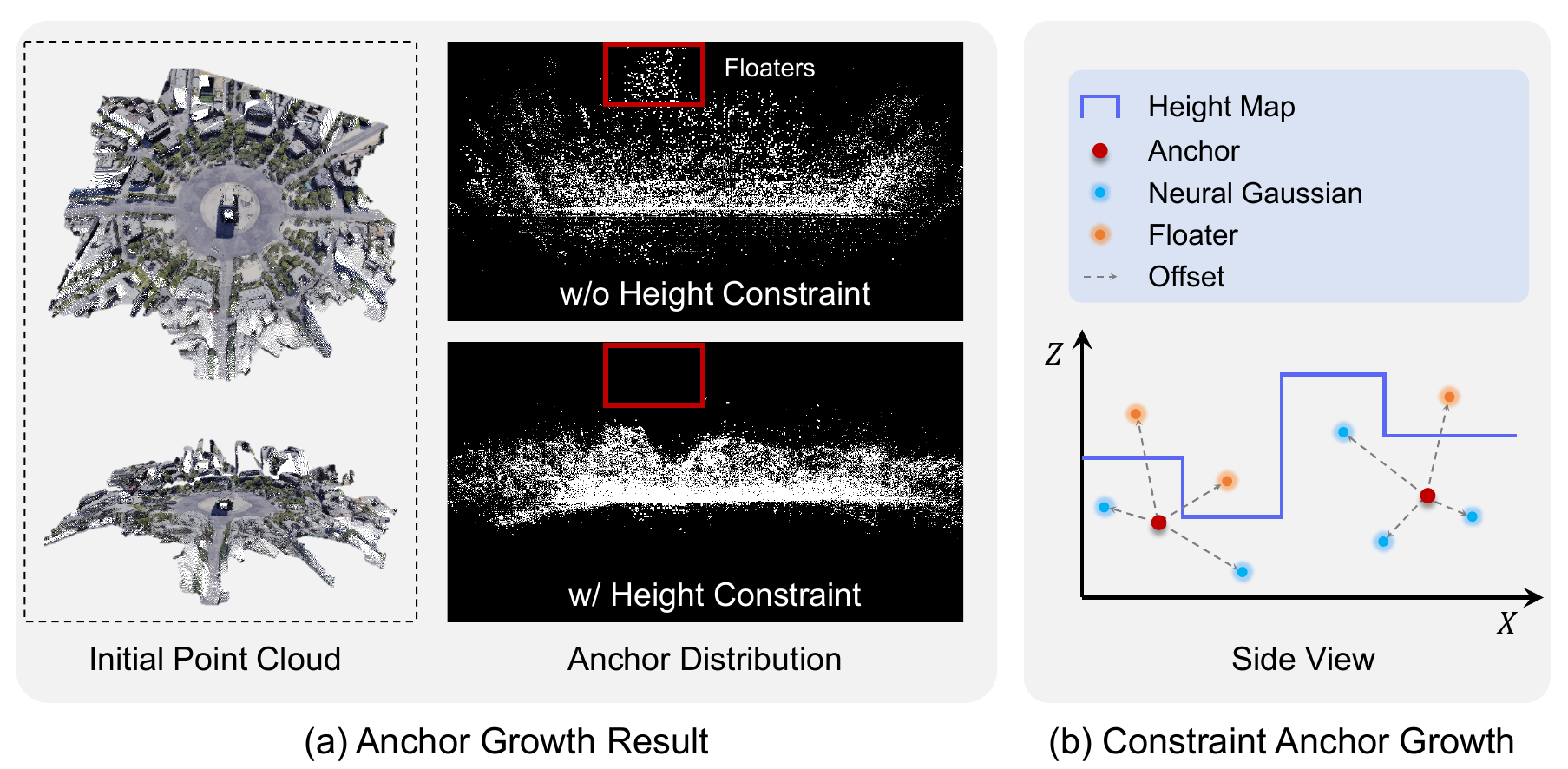}
    \caption{Illustration of anchor growth. (a) Anchor distribution after growth. The original anchors contain numerous floating artifacts and deviate significantly from the initial point cloud, while the Gaussians become more compact and precise with the constraint. (b) The proposed height-constrained anchor growth strategy, where only neural Gaussians whose heights fall within the predefined bounds are considered as candidate new anchors.}
    \label{fig:grow}
\end{figure}

During optimization, the model computes the average gradient of neural Gaussians within each local spatial neighborhood, and regions with large gradients are identified as important areas where new anchors are added. Our method specifically aims to identify potential floater Gaussians and prevent wrong anchor growth in these regions. A key characteristic in remote sensing scenarios is that many aerial images are captured from a nadir viewpoint, where a point’s height is negatively correlated with its distance to the camera. This property allows us to detect Gaussians that are anomalously close to the camera based on their height. Therefore, we construct a coarse height map from the initial dense point cloud to identify Gaussians that may cause camera-oriented expansion. Instead of enforcing a strict geometric supervision, the height map serves as a soft criterion to prevent unreasonable anchor growth while preserving valid scene structures. Specifically, we partition the XY plane into grids and estimate a relaxed upper bound on height within each grid, which is used to suppress anchor growth only for Gaussians that clearly fall outside the valid height as shown in Fig.~\ref{fig:grow}(b). The height map is formulated as:
\begin{equation}
    \mathcal{H}(x,y)=\rho\left(\{z_n\mid {P}_n\in\mathcal{P}_{i,j},\left\lfloor\frac{x_n}{s}\right\rfloor=i,\left\lfloor\frac{y_n}{s}\right\rfloor=j\}\right)
\end{equation}

\noindent where $\rho(\cdot)$ denotes the height aggregation operator, $\lfloor\cdot\rfloor$ indicates the floor operation, and $s$ represents the grid size.

Based on this design, the new anchor growth criterion can be defined as:

\begin{equation}
    \mathbb{I}_{\mathrm{grow}}(P)=
\begin{cases}
1, & {\text{if}}\ \nabla_{g}>\tau_{g} \wedge  z \leq \mathcal{H}(x,y) \\
0, & \text{otherwise.}
\end{cases}
\end{equation}

\noindent where $\nabla_{g}$ denotes the averaged gradient and $\tau_{g}$ is a pre-defined threshold.

The proposed regularization reduces floating artifacts and encourages a more compact Gaussian distribution, as illustrated in Fig.~\ref{fig:grow}(a).

\subsection{Pseudo View Supervision through Progressive DIBR}

Another critical challenge in sparse-view novel view synthesis is the lack of effective supervision for unobserved regions. Due to limited viewpoints and occlusions, some regions in novel views may not be covered by any input observations, resulting in insufficient Gaussian optimization and severe artifacts such as holes or incomplete structures. Prior studies~\cite{kerbl20233d, xu2022point, jung2024relaxing} have shown that Gaussians struggle to model areas beyond the boundaries defined by the initial point cloud unless they are explicitly encouraged to expand outward. However, existing methods~\cite{jung2024relaxing} cannot precisely identify true boundary regions and may introduce redundant primitives. 

To address this issue, we propose to synthesize pseudo views through progressive DIBR to incorporate more prior knowledge from pretrained model. The synthesized pseudo views provide complementary supervision signals, encouraging Gaussian optimization to recover missing structures while maintaining multi-view consistency.

We sample intermediate viewpoints along the acquisition trajectory of the input views to generate additional observations between existing cameras. For each target view $j$, we obtain depth map $\mathbf{D}^{\mathrm{pse}}_j$ by performing depth completion on the dense point cloud. Given target view pixel $\mathbf{p}_j$, we compute the corresponding coordinate $\mathbf{p}_{j\rightarrow i}$ in the source view through depth warping:

\begin{equation}
    \mathbf{p}_{j\rightarrow i}=\mathbf{K}_i\mathbf{T}_{j\rightarrow i}\mathbf{D}^{\mathrm{pse}}_{j}(u,v)\mathbf{K}_j^{-1}\mathbf{p}_j,
\end{equation}
\noindent where $\mathbf{T}_{j\rightarrow i}$ denotes the transformation matrix from view $j$ to view $i$.
Then we sample from the source-view images $\mathbf{I}^{\mathrm{gt}}_i$ to synthesize the pseudo-view image $\mathbf{I}^{\mathrm{pse}}_j$:

\begin{equation}
    \mathbf{I}^{\mathrm{pse}}_{j}(\mathbf{p}_j)=\operatorname{Sample}(\mathbf{I}^{\mathrm{gt}}_i;\mathbf{p}_{j\rightarrow i}),
\end{equation}

\noindent where $\operatorname{Sample}(\cdot)$ denotes the pixel sampling operation. Due to spatial occlusions, it is not possible to obtain all pixels of the target view in a single step. We therefore design a progressive synthesis strategy that integrates images from multiple input views. Specifically, we compute the relative distances between the target view and all input views and sort them accordingly. The pseudo view is filled sequentially from near to far using depth-based warping, and any remaining unfilled areas are finally completed using depth warping with padding. Fig.~\ref{fig:abl_pse} illustrates the detailed synthesis process. This progressive strategy follows the intuition that nearby source views generally provide more reliable correspondence with fewer occlusions. By gradually integrating complementary observations, the synthesized views reduce artifacts caused by unreliable warping.

\subsection{Optimization}

We adopt an end-to-end optimization strategy that jointly trains the neural Gaussian representation and its associated modules. The overall objective is formulated as a pixel-wise reconstruction loss. During training, we jointly optimize three groups of parameters: (i) the learnable attributes of anchor Gaussians, (ii) the neural Gaussian decoder MLP, and (iii) the projection and fusion networks for IBR features. Besides, we adopt a scheduled supervision strategy. During the early training stage, only ground truth images guide the model to reconstruct the global scene structure and establish stable geometry and appearance. Then pseudo-view supervision is introduced periodically to enforce multi-view consistency and eliminate holes.  Finally, pseudo-view supervision is disabled to focus on optimizing fine-grained details.

The final loss function can be expressed as:

\begin{equation}
   \mathcal{L}= (1-\lambda)\mathcal{L}_1(\hat{\mathbf{I}}, \mathbf{I}) + \lambda \mathcal{L}_{\text{D-SSIM}}(\hat{\mathbf{I}}, \mathbf{I}),
\end{equation}

\noindent where $\hat{\mathbf{I}}$ is the rendered image and $\mathbf{I}$ is the corresponding ground-truth or pseudo-view image. We set $\lambda =0.2$ in our experiments.

\section{Experiment}
\subsection{Experimental Setup}
\subsubsection{Dataset}

\begin{figure*}
    \centering
\includegraphics[width=0.95\textwidth]{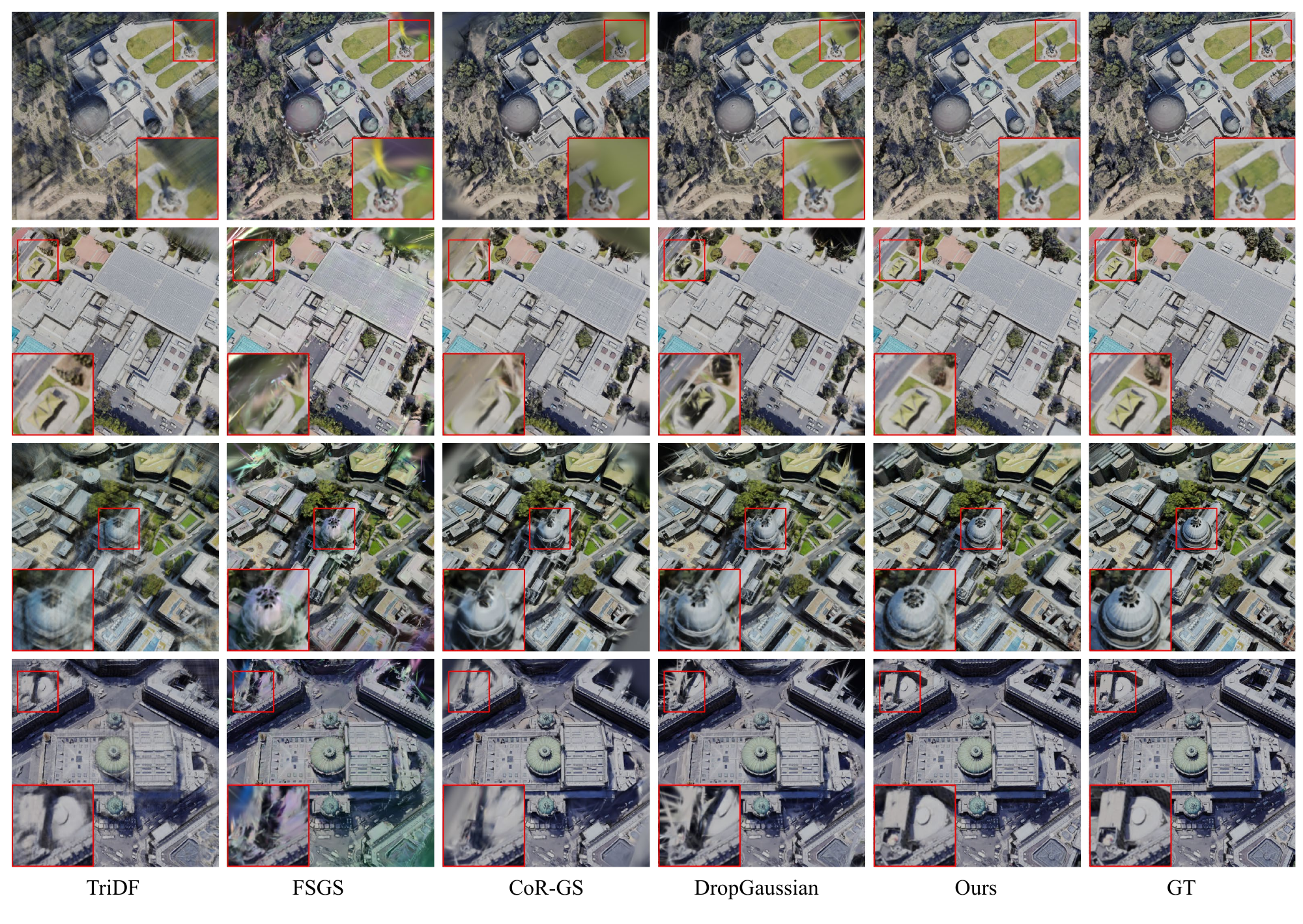}
    \caption{Visual comparison on LEVIR-NVS Dataset.}
    \label{fig:exp}
\end{figure*}

We evaluate our method on LEVIR-NVS~\cite{wu2022remote}, a public remote sensing NVS dataset. It comprises 16 scenes derived from Google Earth, simulating aerial imaging viewpoints and covering diverse land-cover types, including mountainous terrain, urban areas, villages, and building complexes. Each scene contains 21 multi-view images at a resolution of 512$\times$512. To evaluate performance under sparse-view settings, we adopt the data split protocol of TriDF~\cite{kang2025tri}, uniformly selecting 3 images per scene for training and using the remaining views for evaluation.

\subsubsection{Baseline and Metrics}

We compare our proposed method with state-of-the-art approaches, including NeRF-based methods such as RegNeRF~\cite{niemeyer2022regnerf} and FreeNeRF~\cite{yang2023freenerf}, as well as 3DGS–based methods including FSGS~\cite{zhu2024fsgs}, DropGaussian~\cite{Park_2025_CVPR}, CoR-GS~\cite{zhang2024cor} and the vanilla 3DGS~\cite{kerbl20233d}. In addition, we compare against advanced methods specifically designed for remote sensing scenarios, including MPNeRF~\cite{gao2025mp} and TriDF~\cite{kang2025tri}. For all compared methods, we follow the same experimental settings and training protocols to ensure a fair comparison. Following prior work, we report PSNR, SSIM, LPIPS and AVGE metric~\cite{niemeyer2022regnerf} to evaluate the rendering quality.

\subsection{Implementation Details}

\subsubsection{Architecture Details}

We use the frozen Depth Anything-v2~\cite{yang2024depth} model to extract monocular depth maps from the training images, and then invert the disparity to obtain depth. Next, we perform affine transformation using the sparse point clouds estimated by COLMAP~\cite{schonberger2016structure} to obtain dense depth maps with real-world scale. To eliminate overlapping regions in the point cloud, we reproject image pixels into 3D space in the world coordinate system and uniformly partition the XY plane into grids, where we compute the height distribution of the point cloud within each grid. When the significance level of the dip test is below 0.05, we consider the grid to contain overlap. The horizontal and vertical distance thresholds $\varepsilon_{xy}$ and $\varepsilon_{z}$ for redundant point pairs are set to 0.1 and 1.0, respectively. After obtaining the non-overlapping dense point cloud, we perform voxel downsampling with a voxel size of 1.0, and then merge it with the point cloud acquired via SfM for the initialization of neural Gaussians.

To extract reference features, we employ frozen ResNet-50~\cite{he2016deep} and ViT-B/8 vision transformer~\cite{dosovitskiy2020image}, both pretrained with a DINO~\cite{caron2021emerging} objective. We extract feature maps from the first four layers of ResNet-50 and project them into a unified feature space using $1\times1$ convolutions, followed by upsampling to the size of $\frac{H}{2}\times\frac{W}{2}\times C_r$. The output of ViT includes both global and local tokens, which are projected into the same feature space using linear layers. To ensure consistency with the size of the ResNet feature maps, these tokens are replicated spatially. Finally, all feature maps are summed together to generate the reference feature map with a dimension $C_r=32$. 

For the neural Gaussians, the grid size of the anchors is set to 0.03, with each anchor corresponding to $k = 10$ neural Gaussians. The modulation network $\mathcal{E}_\mathrm{mod}$ consists of linear layers followed by a Sigmoid activation function, with initialization settings designed to produce small output values, thereby stabilizing the optimization process. The dimension of the anchor features is set to 32.

We construct a height map using the dense point cloud to constrain the growth of neural Gaussians and suppress artifacts. The grid size for partitioning the XY plane is set to 10. We compute the 95th percentile of the height within each grid, which is then multiplied by a scaling factor to relax the boundary. To avoid disrupting fine structural details, we apply the height constraint only at the lowest resolution of the anchor voxels. To avoid excessive deviation from the training viewpoints, we randomly sample pseudo-view camera poses from the given circular acquisition trajectory.

\begin{table}[]
    \centering
    \caption[toc entry]{Quantitative comparison of rendering quality and efficiency between different methods. 
     The best, second-best, and third-best entries are marked in     
   \begin{tikzpicture}
        \draw[fill=bestcolor, draw=white] (0,0) rectangle (0.6,0.25);
    \end{tikzpicture}, 
    \begin{tikzpicture}
        \draw[fill=secondcolor, draw=white] (0,0) rectangle (0.6,0.25);
    \end{tikzpicture}, and 
    \begin{tikzpicture}
        \draw[fill=thirdcolor, draw=white] (0,0) rectangle (0.6,0.25);
    \end{tikzpicture}, 
    respectively.
    }
     \setlength{\extrarowheight}{1pt}
    \resizebox{0.5\textwidth}{!}{
    \begin{tabular}{l|cccc|c}
    \toprule
        Method  & PSNR$\uparrow$ & SSIM$\uparrow$ & LPIPS$\downarrow$ & AVGE$\downarrow$ & FPS$\uparrow$ \\
    \midrule
    RegNeRF~\cite{niemeyer2022regnerf} & 19.83 & 0.695 & 0.389 & 0.131 & 0.19 \\
FreeNeRF~\cite{yang2023freenerf} & 19.04 & 0.524 & 0.373 & 0.148 & 0.19 \\
MPNeRF ~\cite{gao2025mp} & 21.72 & \cellcolor{thirdcolor}0.800 & \cellcolor{secondcolor}0.190 & 0.083 & 0.18 \\
TriDF~\cite{kang2025tri} & \cellcolor{secondcolor}24.07 & \cellcolor{secondcolor}0.820 & 0.213 & \cellcolor{secondcolor}0.071 & 0.20 \\
\midrule
3DGS~\cite{kerbl20233d} & 18.29 & 0.593 & 0.313 & 0.144 & 280 \\
FSGS~\cite{zhu2024fsgs} & 21.18 & 0.772 & 0.230 & 0.094 & 343 \\
CoR-GS~\cite{zhang2024cor} & \cellcolor{thirdcolor}22.39 & 0.793 & 0.212 & \cellcolor{thirdcolor}0.082 & 268 \\
DropGaussian~\cite{Park_2025_CVPR} & 21.89 & 0.793 & \cellcolor{thirdcolor}0.201 & 0.084 & 282 \\
\textbf{Ours} & \cellcolor{bestcolor}30.90 & \cellcolor{bestcolor}0.938 & \cellcolor{bestcolor}0.076 & \cellcolor{bestcolor}0.025 & 177 \\
    \bottomrule
    \end{tabular}
    }
    \label{tab:result_comparison}
\end{table}

\subsubsection{Training Details}

The neural Gaussian anchor attributes $\mathcal{O}_{k}$, $l_v$, decoder MLP $F_{\{\alpha,q,s,c\}}$, feature projection network $\mathcal{E}_\mathrm{ref}$, and modulation network $\mathcal{E}_\mathrm{mod}$ are jointly optimized in an end-to-end manner. We use the Adam optimizer, with different initial learning rates assigned to different network modules, and apply a continuous exponential learning rate scheduling strategy. The total number of training iterations is 30,000. The projection module $\mathcal{P}$ associated with the ResNet and ViT features is trained jointly with the neural Gaussian model using the AdamW optimizer, with an initial learning rate of $1\times10^{-4}$ and its parameters are frozen after 20,000 iterations. Neural Gaussian densification is performed every 100 iterations from iteration 1,500 to 15,000. In addition, pseudo-view sampling is introduced for training once every 50 iterations from iteration 3,000 to 20,000. All experiments are conducted on a single Nvidia GeForce RTX 4090 GPU.

\begin{table*}[htbp]
  \centering
  \caption{Quantitative comparison per scene on LEVIR-NVS dataset.}
  \begin{adjustbox}{width=\linewidth}
    \begin{tabular}{c|c|ccccccccccccccccc}
    \toprule
    \toprule
    \multicolumn{1}{c}{\rotatebox{60}{Metrics}} & \multicolumn{1}{c}{\rotatebox{60}{Methods}} & \rotatebox{60}{Building\#1} & \rotatebox{60}{Church} & \rotatebox{60}{College} & \rotatebox{60}{Mountain\#1} & \rotatebox{60}{Mountain\#2} & \rotatebox{60}{Observation} & \rotatebox{60}{Building\#2} & \rotatebox{60}{Town\#1} & \rotatebox{60}{Stadium} & \rotatebox{60}{Town\#2} & \rotatebox{60}{Mountain\#3} & \rotatebox{60}{Town\#3} & \rotatebox{60}{Factory} & \rotatebox{60}{Park} & \rotatebox{60}{School} & \rotatebox{60}{Downtown} & \rotatebox{60}{Mean} \\
    \midrule
    \multirow{9}[0]{*}{PSNR}   
          & RegNeRF~\cite{niemeyer2022regnerf} & 13.52 & 14.74 & 17.63 & 20.88 & 20.57 & 15.24 & 14.81 & 20.84 & 22.35 & 21.15 & 26.50 & 21.20 & 23.49 & \cellcolor{thirdcolor}24.74 & 20.11 & 19.46 & 19.83 \\
          & FreeNeRF~\cite{yang2023freenerf} & 15.84 & 15.99 & 19.74 & 19.64 & 21.41 & 13.76 & 15.85 & 20.58 & 23.41 & 14.66 & 20.56 & 19.05 & 21.03 & 23.84 & \cellcolor{thirdcolor}21.81 & 17.46 & 19.04 \\
          & MPNeRF~\cite{gao2025mp} & \cellcolor{thirdcolor}18.81 & 17.93 & 20.71 & 25.50 & 24.92 & 19.56 & 18.64 & 21.59 & 22.08 & 21.20 & 28.57 & \cellcolor{thirdcolor}21.41 & 22.61 & 23.57 & 20.71 & 19.73 & 21.72 \\
          & TriDF~\cite{kang2025tri} & \cellcolor{secondcolor}21.42 & \cellcolor{secondcolor}19.56 & \cellcolor{secondcolor}22.98 & \cellcolor{secondcolor}27.70 & \cellcolor{secondcolor}27.05 & \cellcolor{secondcolor}20.70 & \cellcolor{secondcolor}20.74 & \cellcolor{secondcolor}24.47 & \cellcolor{secondcolor}25.45 & \cellcolor{secondcolor}22.36 & \cellcolor{secondcolor}30.70 & \cellcolor{secondcolor}22.86 & \cellcolor{secondcolor}25.28 & \cellcolor{secondcolor}27.62 & \cellcolor{secondcolor}24.02 & \cellcolor{secondcolor}22.19 & \cellcolor{secondcolor}24.07 \\
          & 3DGS~\cite{kerbl20233d} & 16.27 & 15.16 & 16.27 & 20.16 & 21.58 & 16.68 & 16.73 & 18.54 & 18.46 & 17.22 & 24.21 & 17.93 & 20.25 & 20.64 & 16.49 & 16.04 & 18.29 \\
          & FSGS~\cite{zhu2024fsgs} & 17.69 & 17.52 & 20.01 & 25.03 & 24.39 & 19.76 & 18.36 & 20.74 & 23.42 & 19.92 & 26.46 & 21.11 & 22.47 & 23.93 & 18.42 & 19.63 & 21.18 \\
          & CoR-GS~\cite{zhang2024cor} & 18.39 & \cellcolor{thirdcolor}18.13 & 20.57 & \cellcolor{thirdcolor}26.30 & \cellcolor{thirdcolor}25.68 & \cellcolor{thirdcolor}20.44 & \cellcolor{thirdcolor}20.09 & \cellcolor{thirdcolor}22.64 & \cellcolor{thirdcolor}23.83 & \cellcolor{thirdcolor}21.58 & \cellcolor{thirdcolor}29.41 & 21.26 & \cellcolor{thirdcolor}23.72 & 24.37 & 21.32 & \cellcolor{thirdcolor}20.49 & \cellcolor{thirdcolor}22.39 \\
          & DropGaussian~\cite{Park_2025_CVPR} & 17.94 & 17.67 & \cellcolor{thirdcolor}20.78 & 26.01 & 25.53 & 20.25 & 19.08 & 21.64 & 23.30 & 20.75 & 29.22 & 20.91 & 23.10 & 23.85 & 20.48 & 19.87 & 21.90 \\
          & Ours & \cellcolor{bestcolor}27.38 & \cellcolor{bestcolor}26.41 & \cellcolor{bestcolor}29.75 & \cellcolor{bestcolor}33.96 & \cellcolor{bestcolor}34.50 & \cellcolor{bestcolor}26.83 & \cellcolor{bestcolor}29.30 & \cellcolor{bestcolor}31.22 & \cellcolor{bestcolor}31.07 & \cellcolor{bestcolor}29.71 & \cellcolor{bestcolor}36.83 & \cellcolor{bestcolor}31.04 & \cellcolor{bestcolor}34.24 & \cellcolor{bestcolor}32.97 & \cellcolor{bestcolor}29.37 & \cellcolor{bestcolor}29.89 & \cellcolor{bestcolor}30.90 \\
    \midrule
    \multirow{9}[0]{*}{SSIM} 
          & RegNeRF~\cite{niemeyer2022regnerf} & 0.390 & 0.495 & 0.580 & 0.680 & 0.705 & 0.457 & 0.410 & 0.756 & 0.844 & 0.836 & 0.851 & 0.781 & \cellcolor{thirdcolor}0.897 & \cellcolor{thirdcolor}0.884 & 0.782 & 0.745 & 0.695 \\
          & FreeNeRF~\cite{yang2023freenerf} & 0.441 & 0.482 & 0.541 & 0.414 & 0.511 & 0.140 & 0.378 & 0.627 & 0.788 & 0.331 & 0.294 & 0.643 & 0.714 & 0.810 & 0.692 & 0.584 & 0.524 \\
          & MPNeRF~\cite{gao2025mp} & \cellcolor{thirdcolor}0.730 & \cellcolor{secondcolor}0.720 & \cellcolor{secondcolor}0.790 & 0.820 & 0.810 & \cellcolor{secondcolor}0.730 & 0.710 & 0.810 & 0.800 & \cellcolor{thirdcolor}0.840 & \cellcolor{secondcolor}0.890 & \cellcolor{secondcolor}0.840 & 0.860 & 0.850 & \cellcolor{thirdcolor}0.790 & 0.760 & \cellcolor{thirdcolor}0.800  \\
          & TriDF~\cite{kang2025tri} & \cellcolor{secondcolor}0.825 & \cellcolor{thirdcolor}0.693 & \cellcolor{thirdcolor}0.759 & 0.796 & 0.839 & 0.660 & \cellcolor{secondcolor}0.756 & \cellcolor{secondcolor}0.847 & \cellcolor{secondcolor}0.882 & \cellcolor{secondcolor}0.848 & \cellcolor{thirdcolor}0.886 & \cellcolor{thirdcolor}0.822 & \cellcolor{secondcolor}0.913 & \cellcolor{secondcolor}0.918 & \cellcolor{secondcolor}0.857 & \cellcolor{secondcolor}0.813 & \cellcolor{secondcolor}0.820 \\
          & 3DGS~\cite{kerbl20233d} & 0.622 & 0.460 & 0.385 & 0.447 & 0.687 & 0.479 & 0.521 & 0.626 & 0.646 & 0.617 & 0.694 & 0.606 & 0.857 & 0.846 & 0.478 & 0.524 & 0.593 \\
          & FSGS~\cite{zhu2024fsgs} & 0.696 & 0.642 & 0.682 & 0.813 & 0.840 & 0.702 & 0.703 & 0.795 & 0.837 & 0.798 & 0.864 & 0.806 & 0.870 & 0.863 & 0.708 & 0.732 & 0.772 \\
          & CoR-GS~\cite{zhang2024cor} & 0.707 & 0.665 & 0.693 & \cellcolor{thirdcolor}0.825 & \cellcolor{thirdcolor}0.843 & 0.705 & \cellcolor{thirdcolor}0.749 & 0.810 & \cellcolor{thirdcolor}0.857 & 0.824 & 0.883 & 0.816 & 0.885 & 0.860 & 0.781 & \cellcolor{thirdcolor}0.792 & 0.793 \\
          & DropGaussian~\cite{Park_2025_CVPR} & 0.718 & 0.668 & 0.707 & \cellcolor{secondcolor}0.834 & \cellcolor{secondcolor}0.852 & \cellcolor{thirdcolor}0.712 & 0.739 & \cellcolor{thirdcolor}0.814 & 0.846 & 0.819 & 0.882 & 0.807 & 0.874 & 0.871 & 0.774 & 0.776 & 0.793 \\
          & Ours & \cellcolor{bestcolor}0.935 & \cellcolor{bestcolor}0.893 & \cellcolor{bestcolor}0.914 & \cellcolor{bestcolor}0.931 & \cellcolor{bestcolor}0.953 & \cellcolor{bestcolor}0.886 & \cellcolor{bestcolor}0.940 & \cellcolor{bestcolor}0.949 & \cellcolor{bestcolor}0.949 & \cellcolor{bestcolor}0.942 & \cellcolor{bestcolor}0.955 & \cellcolor{bestcolor}0.947 & \cellcolor{bestcolor}0.976 & \cellcolor{bestcolor}0.968 & \cellcolor{bestcolor}0.934 & \cellcolor{bestcolor}0.942 & \cellcolor{bestcolor}0.938 \\
    \midrule
    \multirow{9}[0]{*}{LPIPS} 
          & RegNeRF~\cite{niemeyer2022regnerf} & 0.568 & 0.504 & 0.482 & 0.468 & 0.445 & 0.509 & 0.554 & 0.324 & 0.268 & 0.261 & 0.343 & 0.333 & 0.218 & 0.242 & 0.364 & 0.379 & 0.389 \\
          & FreeNeRF~\cite{yang2023freenerf} & 0.445 & 0.398 & 0.354 & 0.417 & 0.401 & 0.543 & 0.439 & 0.318 & 0.221 & 0.454 & 0.494 & 0.299 & 0.257 & 0.233 & 0.305 & 0.386 & 0.373 \\
          & MPNeRF & \cellcolor{thirdcolor}0.210 & \cellcolor{secondcolor}0.240 & \cellcolor{secondcolor}0.180 & \cellcolor{thirdcolor}0.200 & \cellcolor{secondcolor}0.180 & \cellcolor{thirdcolor}0.250 & 0.240 & \cellcolor{thirdcolor}0.180 & 0.200 & \cellcolor{secondcolor}0.160 & \cellcolor{secondcolor}0.120 & \cellcolor{secondcolor}0.170 & 0.120 & 0.140 & \cellcolor{thirdcolor}0.200 & \cellcolor{secondcolor}0.190 & \cellcolor{secondcolor}0.190 \\
          & TriDF~\cite{kang2025tri} & \cellcolor{secondcolor}0.188 & 0.322 & \cellcolor{thirdcolor}0.264 & 0.268 & 0.218 & 0.339 & 0.263 & 0.188 & \cellcolor{secondcolor}0.138 & \cellcolor{thirdcolor}0.170 & 0.190 & 0.223 & \cellcolor{secondcolor}0.110 & \cellcolor{secondcolor}0.114 & \cellcolor{secondcolor}0.170 & 0.241 & 0.213 \\
          & 3DGS~\cite{kerbl20233d} & 0.282 & 0.393 & 0.450 & 0.433 & 0.267 & 0.365 & 0.359 & 0.281 & 0.277 & 0.296 & 0.287 & 0.317 & \cellcolor{thirdcolor}0.118 & \cellcolor{thirdcolor}0.131 & 0.397 & 0.363 & 0.313 \\
          & FSGS~\cite{zhu2024fsgs} & 0.264 & 0.288 & 0.281 & 0.225 & 0.214 & 0.260 & 0.272 & 0.189 & 0.180 & 0.195 & 0.195 & 0.194 & 0.147 & 0.165 & 0.322 & 0.286 & 0.230 \\
          & CoR-GS~\cite{zhang2024cor} & 0.247 & 0.272 & 0.295 & 0.226 & 0.206 & 0.264 & \cellcolor{thirdcolor}0.234 & 0.190 & 0.161 & 0.191 & \cellcolor{thirdcolor}0.173 & \cellcolor{thirdcolor}0.188 & 0.145 & 0.169 & 0.211 & 0.215 & 0.212 \\
          & DropGaussian~\cite{Park_2025_CVPR} & 0.232 & \cellcolor{thirdcolor}0.260 & 0.270 & \cellcolor{secondcolor}0.200 & \cellcolor{thirdcolor}0.193 & \cellcolor{secondcolor}0.246 & \cellcolor{secondcolor}0.222 & \cellcolor{secondcolor}0.177 & \cellcolor{thirdcolor}0.159 & 0.182 & 0.176 & 0.189 & 0.149 & 0.153 & \cellcolor{thirdcolor}0.200 & \cellcolor{thirdcolor}0.214 & \cellcolor{thirdcolor}0.201 \\
          & Ours & \cellcolor{bestcolor}0.074 & \cellcolor{bestcolor}0.125 & \cellcolor{bestcolor}0.101 & \cellcolor{bestcolor}0.090 & \cellcolor{bestcolor}0.068 & \cellcolor{bestcolor}0.127 & \cellcolor{bestcolor}0.070 & \cellcolor{bestcolor}0.069 & \cellcolor{bestcolor}0.068 & \cellcolor{bestcolor}0.073 & \cellcolor{bestcolor}0.065 & \cellcolor{bestcolor}0.067 & \cellcolor{bestcolor}0.031 & \cellcolor{bestcolor}0.045 & \cellcolor{bestcolor}0.075 & \cellcolor{bestcolor}0.074 & \cellcolor{bestcolor}0.076 \\
    \midrule
    \midrule
    \end{tabular}%
  \end{adjustbox}
  \label{tab:detailed_result}
\end{table*}

\begin{figure*}
    \centering
    \includegraphics[width=\textwidth]{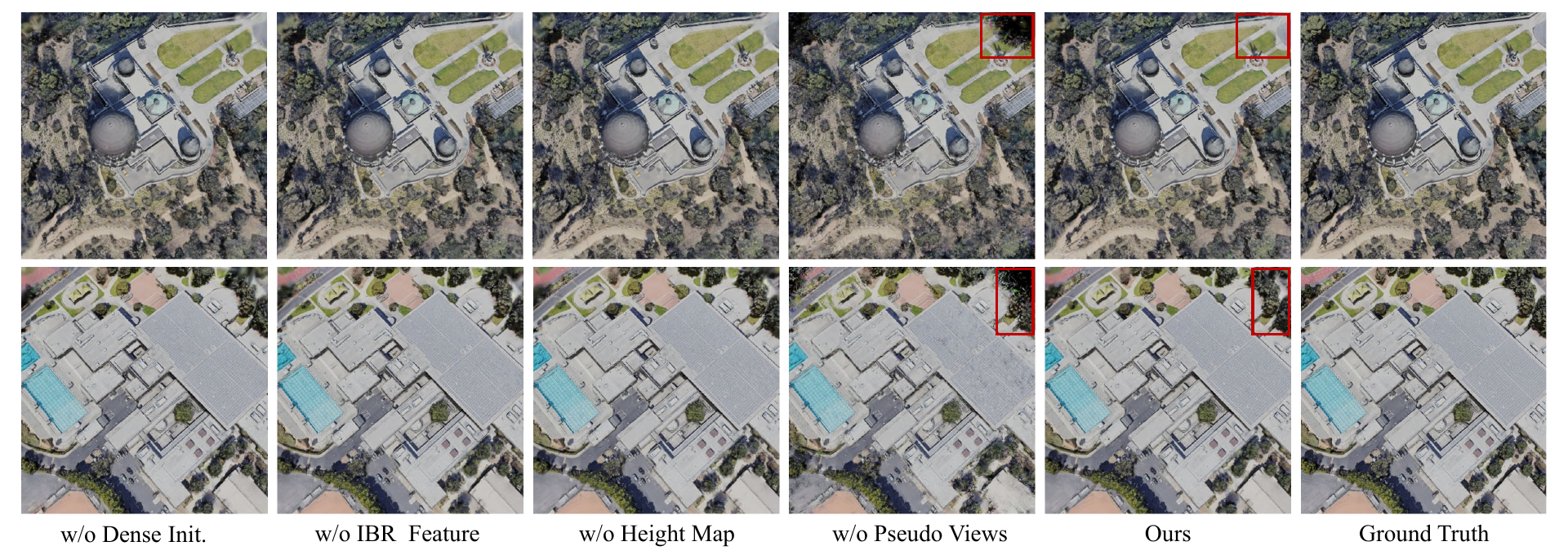}
    \caption{Visualization of rendering quality when removing  different model components.}
    \label{fig:abl_arch}
\end{figure*}

\begin{table}[t]
\centering
\caption{Ablation study of different components. The AVGE values are multiplied by 1e2.}
\label{tab:abl_architecture}
\begin{tabular}{l|cccc}
\toprule
Setting & PSNR$\uparrow$ & SSIM$\uparrow$ & LPIPS$\downarrow$ & AVGE$\downarrow$ \\
\midrule
w/o Dense Init.   & 29.18 & 0.917 & 0.097 & 3.232 \\
w/o IBR Feature   & 30.56 & 0.933 & 0.082 & 2.652 \\
w/o Height Map    & 30.68 & 0.935 & 0.080 & 2.593 \\
w/o Pseudo Views  & 21.06 & 0.782 & 0.186 & 8.795 \\
Ours              & \textbf{30.90} & \textbf{0.938} & \textbf{0.076} & \textbf{2.488} \\
\bottomrule
\end{tabular}
\end{table}

\subsection{Comparison with Other Methods}

The quantitative results on the LEVIR-NVS dataset are reported in Table~\ref{tab:result_comparison}. Experimental results demonstrate that our method achieves superior reconstruction quality over all competing approaches, encompassing both advanced remote-sensing methods and sparse-view 3DGS-based methods. 
We also report the per-scene experimental results, as shown in  Table~\ref{tab:detailed_result}. These results demonstrate that our approach delivers consistently stable, high-quality rendering across scenes of different types and complexities.

Fig.~\ref{fig:exp} presents visual comparisons on several representative scenes. Due to the inherent complexity of remote sensing scenes and the extremely sparse input views, existing methods often suffer from incomplete structures, blurred boundaries, and inconsistent details. While TriDF preserves the overall scene structure, it still suffers from blurred textures and insufficient detail recovery in local regions. FSGS can reconstruct coarse geometric structures but shows noticeable artifacts such as ghosting and shape distortion in challenging areas. CoR-GS and DropGaussian alleviate certain artifacts but still exhibit issues including blurred edges and missing details. In contrast, the proposed method consistently produces sharper structures and recovers more textural details across all scenes. 
This improvement benefits from the unified exploitation of geometric and appearance priors. The geometry-aware initialization with dense depth provides more stable anchors for Gaussian optimization, while cross-view IBR features complement scene representation under limited observations. More importantly, the progressive DIBR-based supervision introduces additional geometric and appearance constraints, enabling the model to recover structures in regions with insufficient observations. As highlighted in the zoomed-in regions, the proposed method maintains more consistent structures and appearance in challenging areas such as image boundaries. Furthermore, it reduces floating artifacts and unreasonable geometric expansion observed in other methods. This demonstrates that the proposed DIBR-based supervision effectively improves multi-view consistency, while the height-constrained anchor growth strategy further stabilizes Gaussian growth. Together, they enhance both structural completeness and rendering stability under sparse-view remote sensing scenes.

\subsection{Ablation Study}

\subsubsection{Ablation of Model Architecture}

To validate the contribution of each component within the overall framework, we conduct an ablation study on the key modules of our model, with the results reported in Table~\ref{tab:abl_architecture} and Fig.~\ref{fig:abl_arch}. Starting from the full model, we individually remove the dense depth initialization, IBR feature fusion, anchor growth with height map constraint, and pseudo-view supervision modules for comparison. Overall, removing any single component leads to varying degrees of performance degradation, demonstrating that these modules play complementary and synergistic roles within the framework. Specifically, removing the dense depth initialization results in noticeable declines in structural consistency and geometric stability, indicating that high-quality initial point clouds are crucial for stable optimization of neural Gaussian under sparse-view settings. Eliminating the IBR feature fusion module degrades texture fidelity and cross-view appearance consistency, suggesting that multi-view feature aggregation effectively compensates for insufficient input observations. The height-constrained anchor growth strategy shows relatively limited influence on quantitative metrics, but it effectively reduces floating artifacts and unreasonable Gaussian expansion, especially in weakly constrained regions. In contrast, removing pseudo-view supervision causes significant performance degradation, particularly in unobserved regions where holes and structural collapse are more likely to occur as shown in Fig.~\ref{fig:abl_arch}. This confirms that pseudo views provide essential multi-view consistency priors. In conclusion, the proposed modules enhance the model from the perspectives of initialization, anchor growth control, appearance enrichment, and multi-view consistency, jointly enabling stable and high-quality results in sparse-view remote sensing scenarios.

\begin{figure*}
    \centering
    \includegraphics[width=\textwidth]{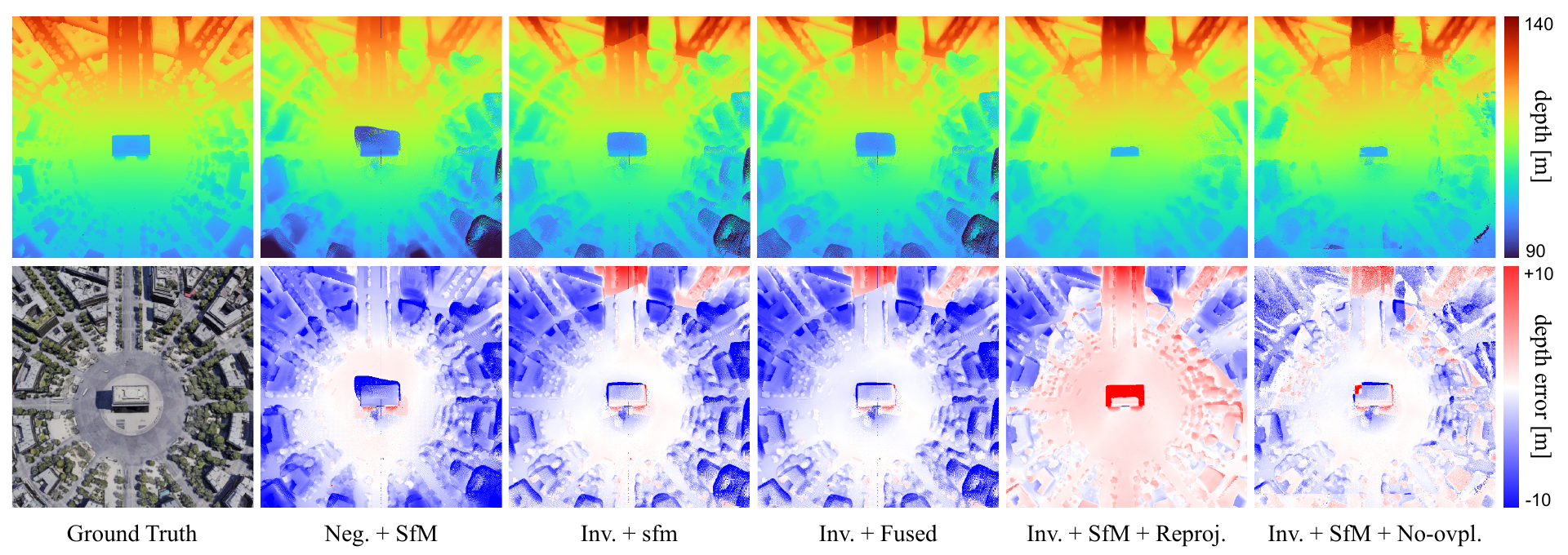}
    \caption{Visualization of depth map and depth error with different initialization strategies.}
    \label{fig:abl_init_map}
\end{figure*}

\subsubsection{Ablation of Initialization}

\begin{table}[t]
\centering
\caption{Quantitative comparison under different mapping, rescaling, and refinement strategies. The AVGE values are multiplied by 1e2.}
\label{tab:abl_init_map}
\begin{tabular}{l l l | c c | c}
\toprule
Map & Source & Refine & MAE$\downarrow$ & Abs Rel$\downarrow$ & AVGE$\downarrow$ \\
\midrule
Neg. & SfM   & --         & 3.72  & 0.0322  & 3.853 \\
Inv. & SfM   & --         & 3.16  & 0.0273  & 3.522 \\
Inv. & Fused & --         & 3.31  & 0.0285  & 3.485 \\
\midrule
Inv. & SfM   & Reproject  & 2.18  & 0.0187  & 3.406 \\
Inv. & SfM   & No ovlp.   & \textbf{1.91}  & \textbf{0.0165}  & \textbf{3.302} \\
\bottomrule
\end{tabular}
\end{table}

\begin{figure}[t]
    \centering
    \includegraphics[width=0.5\textwidth]{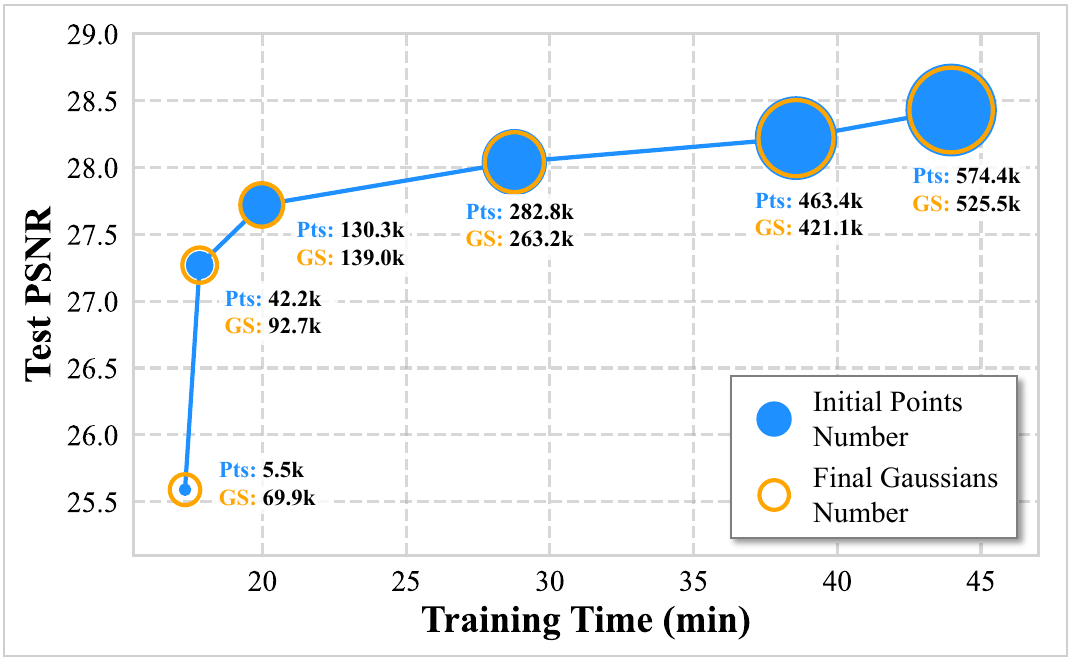}
    \caption{Performance under different size of voxel grid down-sampling.}
    \label{fig:abl_init_number}
\end{figure}
\begin{figure}[t]
    \centering
    \includegraphics[width=0.5\textwidth]{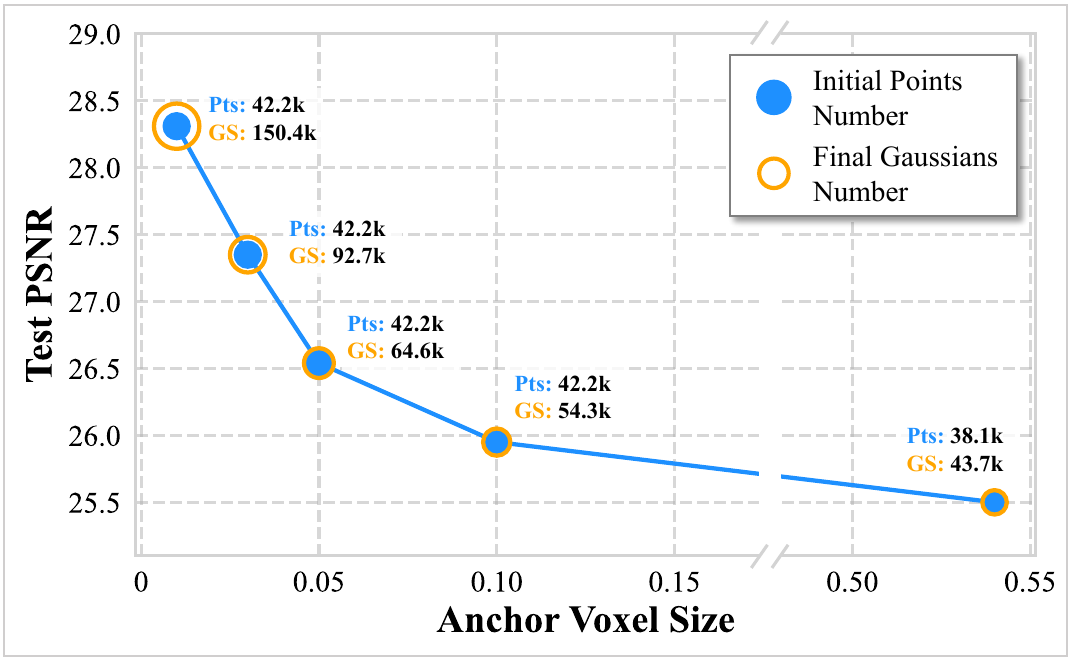}
    \caption{Performance under different size of anchor voxel.}
    \label{fig:abl_init_anchor}
\end{figure}

We evaluate the impact of different initialization strategies on model performance. We assess the quality of the point cloud by measuring the absolute error (MAE and Abs Rel) of the depth map projected from the dense point cloud onto the input views. All depth results are processed using z-buffering. The results are shown in Table~\ref{tab:abl_init_map} and Fig.~\ref{fig:abl_init_map}. We compare different mapping strategies, sources of affine transformation, and point cloud refinement strategies. Among them, Neg. and Inv. represent using the negative or reciprocal values of the original disparity for mapping, respectively. SfM and Fused refer to using the original sparse SfM point cloud and the point cloud processed through Patch Match Stereo densification for affine transformation, respectively. Reproject refers to using multi-view maximum depth to alleviate ambiguities, while No ovlp. is our proposed strategy for removing overlapping point clouds. From the experimental results, it is clear that the reciprocal mapping has higher consistency with the original depth distribution. Although Fused point clouds provide more points for scale adjustment, they also introduce more noise. Additionally, it can be observed that monocular depth tends to underestimate the background depth (indicated by the blue areas in the error map), and this issue becomes more pronounced after multi-view z-buffer processing, as it defaults to retaining the closer depth. The Reproject strategy, which retains the farthest depth from the multi-view images, can alleviate this issue; however, it leads to significant distortion in the foreground depth, which tend to overestimate the depth. Our proposed local point cloud overlap removal strategy effectively mitigates this problem without damaging the foreground depth. We also computed the AVGE metric and found a significant positive correlation with the accuracy of the initialized point cloud, further confirming the importance of the initialization.

\begin{table}[t]
\centering
\caption{Performance comparison for different sizes of voxel grid downsampling.}
\label{tab:abl_init_number}
\begin{tabular}{c|c c|c c c|c}
\toprule
Size & Pts(k) & GS(k) & PSNR$\uparrow$ & SSIM$\uparrow$ & LPIPS$\downarrow$ & T(min)$\downarrow$ \\
\midrule
- & 574.4 & 525.5 & \textbf{28.43} & \textbf{0.954} & \textbf{0.052} & 43.9 \\
0.2 & 463.4 & 421.1 & 28.22 & 0.952 & 0.052 & 38.5 \\
0.3 & 282.8 & 263.2 & 28.04 & 0.950 & 0.056 & 28.7 \\
0.5 & 130.3 & 139.0 & 27.72 & 0.943 & 0.064 & 19.9 \\
1.0 & 42.2 & 92.7 & 27.36 & 0.935 & 0.074 & 17.8 \\
$\infty$ & 5.5 & 69.9 & 25.59 & 0.906 & 0.105 & 17.3 \\
\bottomrule
\end{tabular}
\end{table}

\begin{table}[t]
\centering
\caption{Performance comparison for different sizes of anchor voxel.}
\label{tab:abl_init_anchor}
\begin{tabular}{c|cc|ccc}
\toprule
Voxel Size & Pts(k) & GS(k) &  PSNR$\uparrow$ & SSIM$\uparrow$ & LPIPS$\downarrow$ \\
\midrule
0.01 & 42.2 & 150.4 & \textbf{28.32} & \textbf{0.951} & \textbf{0.057} \\
0.03 & 42.2 & 92.7 & 27.36 & 0.935 & 0.074 \\
0.05 & 42.2 & 64.6 & 26.54 & 0.921 & 0.089 \\
0.10 & 42.2 & 54.3 & 25.95 & 0.909 & 0.100 \\
0.54 & 38.1 & 43.7 & 25.50 & 0.896 & 0.115 \\
\bottomrule
\end{tabular}
\end{table}

Dense depth priors provide the model with initial structural information; however, a trade-off must be made between efficiency and performance, as an excessive amount of redundant point clouds inevitably slows down optimization. To this end, we apply voxel grid downsampling to the dense initial point cloud. We evaluate the impact of different downsampling rates on model performance, and record both the number of initial points and the final number of Gaussian anchors to assess redundancy in the initialization. The results are shown in Fig.~\ref{fig:abl_init_number} and Table~\ref{tab:abl_init_number}, which is conducted on scene Building\#1. As the number of initial points increases, both training time and overall model performance improve accordingly. Nevertheless, a clear elbow point emerges in the curve, beyond which the performance gains become much smaller relative to the increased computational cost. In Fig.~\ref{fig:abl_init_number}, the sizes of the blue and yellow concentric circles represent the numbers of initial points and final Gaussian anchors, respectively. It can be observed that when the initialization contains few points, neural Gaussians must grow extensively to capture scene details. Once the number of initial points exceeds 100k, the final number of Gaussians remains nearly unchanged, indicating that the point number has approached saturation. When the number further exceeds 280k, the final number even decreases, revealing substantial redundancy in the initialization. Overall, downsampling the initial dense point cloud is crucial for improving training efficiency and eliminating redundancy without compromising reconstruction quality.

The size of the anchors reflects the level of granularity used for scene modeling. We evaluate the effect of different anchor sizes on scene Building\#1, with the results shown in Fig.~\ref{fig:abl_init_anchor} and Table~\ref{tab:abl_init_anchor}. With the number of initial points fixed, finer grid resolutions lead to higher reconstruction performance, but at the cost of a larger number of anchors and increased computational overhead.

\subsubsection{Ablation of Network} 
The anchor feature dimension $C_f$ and the IBR reference feature dimension $C_r$ directly affect both the representational capacity and the computational cost of the model. Therefore, we conduct an ablation study on different combinations of dimensions and the results are reported in Table~\ref{tab:feat_channel}.

\begin{table}[t]
\centering
\caption{Effect of different anchor feature and IBR
feature channels.}
\label{tab:feat_channel}
\begin{tabular}{cc|ccc}
\toprule
$C_f$ & $C_r$ & PSNR$\uparrow$ & SSIM$\uparrow$ & LPIPS$\downarrow$ \\
\midrule
32 & 32  & \textbf{30.90} & \textbf{0.938} & \textbf{0.076} \\
32 & 64  & 30.72 & 0.935 & 0.080 \\
32 & 128 & 30.71 & 0.935 & 0.080 \\
32 & 256 & 30.84 & 0.936 & 0.078 \\
\midrule
16 & 64  & 30.39 & 0.930 & 0.088 \\
32 & 64  & 30.72 & 0.935 & 0.080 \\
64 & 64  & 30.85 & 0.937 & 0.077 \\
64 & 128 & 30.70 & 0.935 & 0.079 \\
\bottomrule
\end{tabular}
\end{table}

\begin{table}[t]
\centering
\caption{Ablation study on different network settings.}
\label{tab:abl_net}
\begin{tabular}{l|ccc}
\toprule
Setting & PSNR$\uparrow$ & SSIM$\uparrow$ & LPIPS$\downarrow$ \\
\midrule
w/o MLP $\mathcal{E}_\mathrm{mod}$ & 30.82 & 0.937 & 0.077 \\
w/o ft. $\mathcal{P}$  & 30.79 & 0.936 & 0.079 \\
w/o ResNet Feature  & 30.85 & 0.937 & 0.078 \\
w/o ViT Feature    & 30.81 & 0.936 & 0.078 \\
Ours        & \textbf{30.90} & \textbf{0.938} & \textbf{0.076} \\
\bottomrule
\end{tabular}
\end{table}

When the anchor feature dimension is fixed at 32, 
increasing the IBR feature dimension from 32 to 256 leads to only minor performance variations. In particular, the best performance is achieved when $C_r=32$, while further increasing the IBR feature dimension does not yield noticeable improvements. This suggests that in the current task, IBR features primarily serve to supplement cross-view appearance information, which can be sufficiently represented at relatively low dimensionality. Increasing the number of channels beyond this level introduces redundant information without effectively enhancing the representation capability. We further analyze the influence of the anchor feature dimension. The results show that increasing $C_f$ from 16 to 32 significantly improves model performance, whereas further increasing it to 64 results in only marginal gains. This indicates that, under the complexity of the current scenes, a 32-dimensional feature space is sufficient to effectively encode the key attributes of anchors. Overall, the anchor feature dimension has a more pronounced impact on model performance, while the IBR feature dimension mainly provides auxiliary information. Moreover, we observe that the model tends to achieve better performance when the two feature dimensions are well matched. Considering the trade-off between performance and computational efficiency, we adopt the configuration $C_f = 32$ and $C_r=32$ in the final model, which achieves the best overall performance in our experiments.

We further conduct ablation experiments on the network architecture in Table~\ref{tab:abl_net}. Removing the modulation network or freezing the feature projection module consistently degrades performance, confirming the importance of adaptive IBR feature integration. Excluding either ResNet or ViT features also leads to performance drops, indicating that local texture cues and global semantic information are complementary for cross-view feature modeling.

\begin{figure*}
    \centering
    \includegraphics[width=\textwidth]{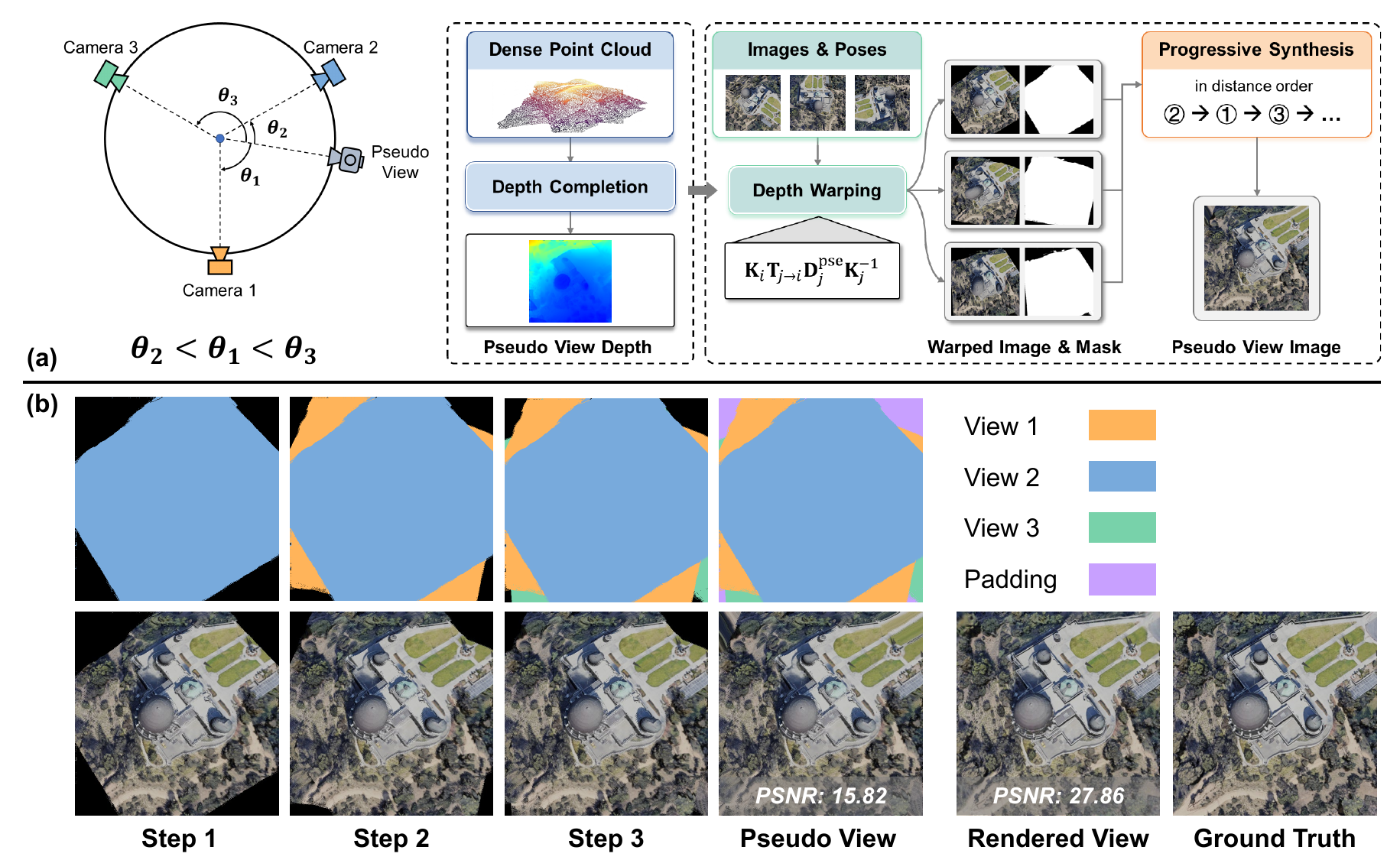}
    \caption{Illustration and analysis of the proposed pseudo view synthesis. (a) Pipeline of the pseudo view synthesis through progressive DIBR.(b) Example of the progressive synthesis process.}
    \label{fig:abl_pse}
\end{figure*}

\subsection{Analysis of DIBR-based Pseudo-view Supervision}

Fig.~\ref{fig:abl_pse}(a) illustrates the pipeline of the proposed progressive DIBR-based supervision strategy. Given the camera trajectory of the training views, intermediate viewpoints are sampled to construct additional supervision signals. For each target viewpoint, its depth map is obtained from the dense point cloud and used to establish geometric correspondence for depth-based warping. Next, images from multiple source views are projected to the target viewpoint via depth-based warping. The source views are integrated progressively according to their distances to the target viewpoint, since closer views generally provide more reliable correspondences with fewer occlusions. In this way, the visible regions of the target view are gradually filled, while the remaining uncovered regions are completed using a padding strategy, resulting in the final pseudo-view image.

Fig.~\ref{fig:abl_pse}(b) provides an example of the progressive synthesis process. The top row visualizes the coverage of different source views in the target view, where the orange, blue, and green regions represent the projected areas from three source views, and the purple regions denote the padding areas. The bottom row shows the corresponding step-by-step synthesis results. As illustrated, the visible regions contributed by different source views vary significantly in the target view, making it difficult for a single view to fully cover the target image. By progressively integrating projections from multiple views, the observable region can be effectively expanded, gradually establishing a more complete supervision signal. In this example, the target pseudo-view is selected from the test set, and we compute the PSNR between the synthesized pseudo-view and the ground-truth image. 

Although the synthesized pseudo-view achieves a relatively low pixel-level fidelity (PSNR = 15.82) due to accumulated depth errors and reprojection inaccuracies, it should be noted that these pseudo views are not expected to provide pixel-perfect targets. Instead, they provide approximate geometric and appearance constraints that complement the sparse original observations. This guidance encourages neural Gaussians to expand into unobserved regions. As a result, the final rendered image produced by the model achieves a PSNR of 27.86. Furthermore, when pseudo-view supervision is removed, the model produces noticeable holes in novel view rendering, as shown in Fig.~\ref{fig:abl_arch}, indicating that supervision from the original views alone is insufficient to cover all visible regions under sparse-view settings. These results demonstrate that the proposed strategy effectively improves supervision coverage and provides valuable multi-view consistency constraints for sparse-view reconstruction.

\section{Conclusion}

This paper addresses the challenge of novel view synthesis in remote sensing scenarios under sparse observations. To this end, we propose DIBR-GS, a neural Gaussian framework that exploits Depth Image-Based Rendering (DIBR) as an additional consistency supervision mechanism between observed and unobserved viewpoints. By integrating reliable geometric initialization, cross-view appearance priors, and progressive DIBR-guided optimization, the proposed framework effectively alleviates geometric ambiguity and insufficient supervision caused by sparse inputs. Extensive experiments demonstrate that DIBR-GS consistently improves reconstruction quality, producing more complete structures, sharper appearance details, and fewer artifacts. These results verify the effectiveness of utilizing DIBR-based supervision for enhancing multi-view consistency in sparse-view remote sensing novel view synthesis. The proposed framework provides a promising solution for remote sensing 3D reconstruction and scene understanding when dense observations are unavailable. However, the proposed method still relies on per-scene optimization. Future research will explore more efficient and generalizable novel view synthesis approaches, such as feed-forward methods.

 % argument is your BibTeX string definitions and bibliography database(s)
%\bibliography{IEEEabrv,../bib/paper}
%

\bibliographystyle{IEEEtran}
\bibliography{IEEEabrv,reference}

\end{document}